\documentclass[11pt]{article}
\usepackage{amsmath}
\usepackage[final]{acl}

\usepackage{times}
\usepackage{latexsym}

\usepackage[T1]{fontenc}

\usepackage[utf8]{inputenc}

\usepackage{microtype}

\usepackage{inconsolata}

\usepackage{graphicx}
\usepackage{algorithm}
\usepackage{algpseudocode}

\usepackage{tikz}
\usetikzlibrary{arrows.meta, positioning}
\usetikzlibrary{arrows.meta, positioning, calc}
\usepackage{amssymb} %

\usepackage{verbatim}

\usepackage{makecell}
\usepackage{tabularx}

\usepackage{adjustbox}
\usepackage{booktabs}
\usepackage[capitalize]{cleveref}
\usepackage{float} %
\usepackage{placeins}
\usepackage{listings}
\usepackage[most]{tcolorbox}
\usepackage{xurl}
\usepackage[disable]{todonotes}

\newcommand{\eh}[1]{}
\newcommand{\og}[1]{}
\newcommand{\ga}[1]{}
\newcommand{\serwar}[1]{}

\title{User-Centric Evidence Ranking for Attribution and Fact Verification}

\author{
  Guy Alt$^{1}$ \quad Eran Hirsch$^{1}$ \quad Serwar Basch$^{2}$ \quad Ido Dagan$^{1}$ \quad Oren Glickman$^{1}$ \\
  $^{1}$Computer Science Department, Bar-Ilan University, Ramat Gan, Israel \\
  $^{2}$UKP Lab, TU Darmstadt, Germany \\
  \texttt{\{guy.alt,eran.hirsch,ido.dagan,oren.glickman\}@biu.ac.il}, \\ \texttt{serwar.basch@tu-darmstadt.de}
}

\begin{document}
\maketitle
\begin{abstract}
Attribution and fact verification are critical challenges in natural language processing for assessing information reliability. While automated systems and Large Language Models (LLMs) aim to retrieve and select concise evidence to support or refute claims, they often present users with either insufficient or overly redundant information, leading to inefficient and error-prone verification. To address this, we propose Evidence Ranking, a novel task that prioritizes presenting sufficient information as early as possible in a ranked list. This minimizes user reading effort while still making all available evidence accessible for sequential verification. We compare two approaches for the new ranking task: one-shot ranking and incremental ranking. We introduce a new evaluation framework, inspired by information retrieval metrics, and construct a unified benchmark by aggregating existing fact verification datasets. Extensive experiments with diverse models show that incremental ranking strategies better capture complementary evidence and that LLM-based methods outperform shallower baselines, while still facing challenges in balancing sufficiency and redundancy. Compared to evidence selection, we conduct a controlled user study  and demonstrate that evidence ranking both reduces reading effort and improves verification. This work provides a foundational step toward  more interpretable, efficient, and user-aligned information verification systems.\footnote{Code and data are available at: \url{https://github.com/guyalt3/User-Centric-Evidence-Ranking-for-Attribution-and-Fact-Verification}}

\end{abstract}

\begin{figure}[t]
\centering
\includegraphics[width=0.99\columnwidth]{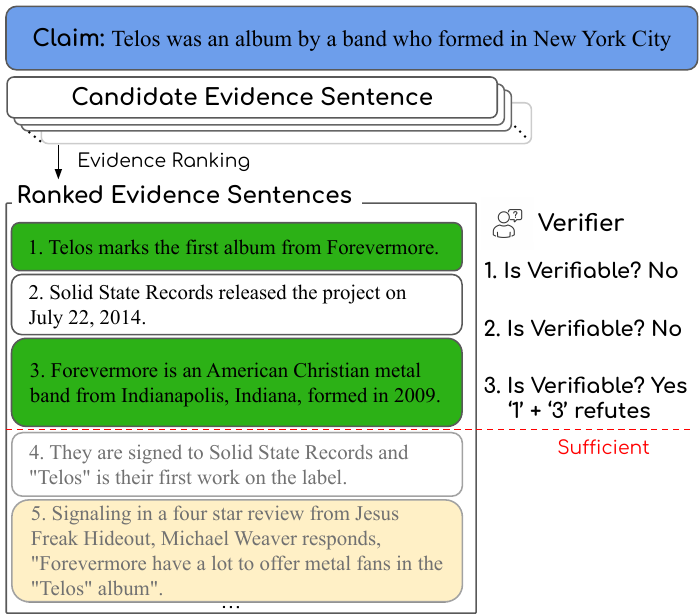}
\caption{Illustration of the \textbf{Evidence Ranking} task. Given a claim and candidate evidence sentences, the system ranks the sentences so that users can stop reading once sufficient evidence is observed. Our evaluation simulates this process by measuring the number of sentences a user would read. 
Green sentences denote relevant evidence, and yellow denotes redundant evidence that can be skipped.}
\label{fig:task_overview}
\end{figure}

\section{Introduction}
\label{sec:introduction}
Large language models (LLMs) can often make claims that appear plausible but are not grounded in reliable evidence, often referred to as \textit{hallucinations} \cite{10.1145/3703155}. Such hallucinations create misinformation problems, making it difficult for users to distinguish between supported and unsupported content \cite{si-etal-2024-large}.
To facilitate verification of their output, LLMs are often instructed to attribute claims with self-citation evidence \citep{gao-etal-2023-enabling}.
Users are then expected to read this evidence to confirm the claims are supported.
A similar situation arises for human fact-checkers, who also rely on automated fact-checking systems to assess claims \cite{nakov2021automatedfactcheckingassistinghuman}.

In both attribution and fact verification settings, the core challenge lies in efficiently identifying a minimal set of evidence that justifies or refutes a claim, without overwhelming the user with irrelevant or redundant information. As illustrated in \cref{fig:task_overview}, there can be many evidence sentences, but Sentences 1 and 3 are sufficient to refute the claim that the album ``Telos'' was by a band formed in New York City. Sentence 1 conveys that ``Telos'' is an album by the band Forevermore, and Sentence 3 clarifies that the band was actually formed in Indianapolis. Replacing Sentence 1 with Sentence 5 would also be considered sufficient, as they convey the same information. A user trying to verify a claim would like to read the minimal number of sentences necessary - Sentences 1 and 3, or Sentences 3 and 5. This exemplifies that evidence sentences exhibit both complementary and redundant characteristics.

Consequently, to simplify the verification process, research has focused on evidence selection systems that output a \emph{minimal} set of evidence \cite{li-etal-2025-minimal, hirsch-etal-2025-laquer, zheng-etal-2024-evidence, atanasova-etal-2022-fact, wan-etal-2021-dqn}.
This formulation naturally leads to evaluating sufficiency using overlap-based metrics such as precision and recall. However, this approach presents a fundamental trade-off: striving for high precision can inadvertently filter out crucial pieces of evidence, while aiming for high recall can result in an overwhelming amount of unnecessary information for the user. 

To overcome this inherent trade-off, we propose a user-aligned formulation of evidence selection that reflects how people verify claims in practice.
In practical verification settings, users typically process information sequentially, discontinuing their search once they have gathered sufficient evidence to either support or refute a claim \citep{hausmann2008sequential}. Following the example in \cref{fig:task_overview}, after reading three sentences, a user would refute the claim that ``Telos was an album by a band who formed in New York City,'' given that the band was formed in Indianapolis. 
Specifically, in both human fact-checking and LLM attribution scenarios, we propose to reframe evidence selection from a binary decision to a ranking task. Rather than selecting a fixed set of evidence, our goal is to generate a global ranking of all candidate sentences, such that sufficient evidence appears as early as possible. 
Importantly, we do not impose any required order among sentences within a sufficient evidence set; instead, success is determined by whether the sentences belonging to a minimal sufficient set are ranked early, regardless of their internal ordering. Overall, this formulation allows users to verify evidence sequentially, reading only as much as needed.

To assess how effectively methods surface informative evidence early while reaching minimal sufficiency, we develop a suite of evaluation metrics inspired by standard Information Retrieval (IR) measures, and construct a unified benchmark from existing fact verification datasets. We investigate a diverse set of modeling approaches, including embedding-based similarity methods, fine-tuned Natural Language Inference (NLI) models, fine-tuned reasoning-based rerankers and LLMs. In addition, we adapt existing evidence selection techniques to an incremental ranking formulation, where each sentence is chosen in the context of previously selected evidence. This encourages a global, non-redundant ranking of candidate sentences.

Our experiments show that incremental strategies generally surface complementary evidence more efficiently than one-shot approaches, and that LLM-based methods achieve the strongest overall performance with an MRR of 0.75. However, our analysis of user reading effort indicates substantial room for improvement. In order to compare our proposed evidence ranking to standard evidence selection, we conduct a controlled user study where users are instructed to verify claims. We show that evidence ranking leads users to read fewer sentences while achieving more accurate verification decisions. Taken together, our formulation, benchmark, metrics, and comparative analysis establish a foundation for evidence ranking, advancing toward more interpretable, efficient, and trustworthy verification systems.

\section{Background}\label{sec:background}
Evidence verification, whether performed by human users or by automated models, presents significant challenges. For instance, users interacting with LLMs often manually inspect multiple source sentences or documents to confirm the validity of the generated claims, which can be time-consuming and error-prone \citep{slobodkin-etal-2024-attribute}. To mitigate this, localized attribution has been proposed to pinpoint relevant evidence and reduce the user's reading burden \citep{radevski2025synthesizing,hirsch-etal-2025-laquer}. Beyond LLM attribution, both automated and human fact verification critically rely on evidence selection methods. These methods aim to identify supporting or refuting evidence, driven by needs for efficiency \citep{jiang-etal-2021-exploring-listwise} and interpretability \cite{paranjape-etal-2020-information}. Critically, despite differing retrieval mechanisms, both attribution and fact verification share a common evidence selection phase: identifying and reasoning over a \textit{sufficient} and \textit{non-redundant} evidence set to validate a claim. The evidence is most frequently segmented at the sentence-level \citep{thorne-etal-2018-fever, kamoi-etal-2023-wice, jiang-etal-2020-hover}.

A common approach for evidence selection is to rerank sentences, then choose from the top-$k$.
Scoring sentences for reranking uses similarity-based over dense embedding representations, or probabilities from fine-tuned NLI models.
While \citet{schuster-etal-2022-stretching, jiang-etal-2021-exploring-listwise} rank each evidence separately, \citet{malviya-katsigiannis-2024-evidence,fajcik-etal-2023-claim} propose hierarchical ranking, suggesting models benefit from selecting complementary information rather than treating all sentences independently. Although ranking was used as an evidence selection strategy, it was not evaluated as a ranking task.
Traditionally, evidence selection is evaluated as a binary task, with standard metrics like precision/recall or F1 scores against predetermined unordered evidence sets.
Such metrics are problematic, as methods returning a single set risk omitting critical information or failing to remove redundant or unnecessary evidence.
Overall, existing evidence selection evaluation does not capture the trade-offs between conciseness and sufficiency.

Other approaches for evidence selection include fine-tuning models for the task \citep{radevski2025synthesizing, paranjape-etal-2020-information}, and zero-shot prompting of LLMs \citep{li-etal-2025-minimal, hirsch-etal-2025-laquer}. Since these approaches do not provide a score for each sentence, they do not naturally apply to the evidence ranking task.

Finally, text reranking has been explored in web search passage reranking \citep{sun-etal-2023-chatgpt,liu2025reasonrankempoweringpassageranking}. While the methods are applicable to our setting, they were not evaluated for facilitating fact verification. Accordingly, our experimental settings includes such methods as a baseline.

\section{Evidence Ranking}
We formulate evidence selection as a ranking problem designed for user verification. The goal is to order all evidence sentences so that a minimal, sufficient set appears first mirroring real-world behavior, where users stop reading once they possess enough information \cite{hausmann2008sequential}. As noted in \cref{sec:background}, we operate on unordered sentence-level evidence sets.

\subsection{Task Definition}\label{sec:task_definition}
Given a claim $c$ and a set of candidate evidence sentences 
$S = \{s_1, \dots, s_n\}$, the goal is to produce a ranking $P = \langle s_{p_1}, \dots, s_{p_n} \rangle$ of $S$.  
In this setting, sentences are examined sequentially according to the ranking.
At any decision point $i$, the accumulated prefix set 
$P_i = \langle s_{p_1}, \dots, s_{p_i} \rangle$, may be sufficient.
A prefix $P_i$ is \emph{sufficient} for claim $c$ if the there exists a subset $E \subseteq P_i$ that confidently verifies or refutes the claim $c$. 

Since multiple sufficient sets may exist, we focus on the smallest prefix reaching sufficiency: the \emph{Minimal Sufficient Rank} (MSR). The objective is to minimize the MSR for each claim, surfacing a minimal sufficient set as early as possible to reduce user effort. Formal definitions are provided in~\cref{app:task-definition}. 

\subsection{Evaluation}             %
\label{sec:evaluation}

As users examine evidence sequentially to decides when they have enough information to confidently assess the claim's veracity \cite{goyal-etal-2023-else}, our metrics evaluate the efficiency of reaching sufficiency. We adapt three established IR metrics to this setting. Our primary metric is a modified \textbf{Mean Reciprocal Rank (MRR)} which quantifies how quickly a ranking reveals a minimal sufficient set (efficiency). As a complementary diagnostic, we also report \textbf{Success Rate (SR)}, which assesses whether sufficiency is achieved exactly at the earliest theoretically optimal position (strict optimality). For completeness, we additionally include an adapted \textbf{Normalized Discounted Cumulative Gain (NDCG)} as a broader measure of ranking quality (see~\cref{app:NDCG-definition} for detailed definition).

\paragraph{Mean Reciprocal Rank (MRR).} 
Building on the well-established Mean Reciprocal Rank (MRR) used in information retrieval \cite{voorhees-tice-2000-trec}, we adapted the metric to evaluate how efficiently a ranking surfaces a minimal sufficient set of evidence. 
MRR is the central metric in our evaluation because it directly aligns with the goal of minimizing the number of sentences a user must read.
In the classic retrieval setting, the ideal reciprocal rank is always $1$, however, in our setting, a single sentence may not be sufficient to verify or refute a claim. As a result, the ideal point of sufficiency may occur later in the ranking, depending on the minimal set required. To account for this variability, we normalize the observed rank by subtracting the Ideal Minimal Sufficient Rank (IMSR), ensuring that scores remain comparable across instances with different sufficiency requirements.  For a formal definition of our MRR see \cref{app:MRR-definition}. 

In the example shown in \cref{fig:task_overview}, the MSR is 3 and the IMSR is $2$, resulting in a reciprocal rank of $\frac{1}{3-2+1}=0.5$.
This value has an intuitive interpretation as a measure of additional reading effort beyond the minimal sufficient set, reflecting user effort in practice. In this specific case, the user reads one additional sentence beyond the minimal sufficient set, calculated as $\frac{1}{0.5}-1=1$.

\paragraph{Success Rate (SR).}
Complementing MRR, we define a stricter criterion inspired by \emph{Success} metrics used in ranking tasks \citep{Matsubara2022EnsembleTF}.
Specifically, Top-1 accuracy reports the proportion of cases where the top-ranked item is correct.
Analogously, our Success Rate (SR) assesses whether a system achieves sufficiency precisely at the empirically optimal position in the ranking. Formally, given the Ideal Minimal Sufficient Rank (IMSR) of a candidate set $S$ for claim $c$, $SR$ is defined as the sufficiency at IMSR. For a formal definition of SR see \cref{app:SR-definition}. 

At the dataset level, mean SR corresponds to the proportion of claims for which the system's ranking achieves sufficiency exactly at the optimal point. This measure also ranges from 0 to 1, with 1 attained only when the system consistently produces an optimal ordering of evidence across all claims.
In the example appearing in \cref{fig:task_overview}, the $SR$ would be 0 as the ranking did not optimally place two sufficient sentences at the top.

\section{Datasets and Benchmark Construction}
\label{sec:datasets}
To evaluate evidence ranking, a benchmark must provide claims alongside candidate and gold evidence sets. As no existing dataset is designed for this sequential, user-focused task, we construct a new benchmark by adapting and unifying three widely-used fact verification resources: FEVER \citep{thorne-etal-2018-fever}, HoVer \citep{jiang-etal-2020-hover}, and WICE \citep{kamoi-etal-2023-wice}. Our goal was to create a diverse testbed that spans the primary challenges in evidence retrieval. Importantly, the gold evidence sets in these datasets are minimal and decontextualized, a property we preserve in our benchmark.

Following our task definition (\cref{sec:task_definition}), any prefix of the ranked candidates that contains an entire gold evidence set is considered \emph{sufficient}. This reframing allows us to measure how quickly different methods surface minimal evidence for verification. 
Next, we describe the characteristics of each dataset, followed by a discussion of our benchmark construction.

\paragraph{FEVER.}  
A large-scale dataset of claims generated through edits of Wikipedia sentences.  
Annotators identified one or more gold evidence sets sufficient for claim verification, with expert review ensuring annotation quality.  
Evidence sets range from single-sentence sufficiency to multi-sentence justifications.  
We exclude claims labeled as \texttt{NOT ENOUGH INFO}, since no gold evidence is available and ranking cannot be meaningfully evaluated.

\begin{table}[!t]
\centering
\Large
\adjustbox{max width=\linewidth}{
    \begin{tabular}{lccc}
    \toprule
    & \textbf{FEVER}
    & \textbf{HoVer}
    & \textbf{WICE} \\
    \midrule
    Number of Instances & 400 & 400 & 200 \\
    Avg. Candidate Evidence Set Size (Sents) & 12.8 & 34.0 & 41.8 \\ 
    Avg. Number of Gold Evidence Sets & 2.1 & 1.0 & 4.9 \\
    Avg. Gold  Set Size (Sents) & 1.4 & 2.8 & 2.8 \\
    Avg. Optimal Gold Evidence Set Size (Sents) & 1.3 & 2.8 & 2.0 \\
    \bottomrule
    \end{tabular}
}
\caption{Overview of our curated datasets, showing the number of instances and key statistics for candidate and gold evidence sets.} 
\label{tab:dataset_sizes}
\end{table}

\paragraph{HoVer.}  
Designed for multi-hop reasoning, HoVer contains claims spanning multiple Wikipedia articles.  
Constructed from extensions of HotpotQA \cite{yang-etal-2018-hotpotqa}, HoVer annotates longer reasoning chains (2-6 hops) that combine multiple sentences into a single gold set.  
The dataset stresses ranking systems to prioritize complementary evidence and handle distributed justification.  

\paragraph{WICE.}  
Derived from Wikipedia, WICE contains naturally occurring claims with candidate evidence retrieved from Common Crawl.
Annotations are performed at the sub-claim level, and gold claim-level sets are constructed by combining one supporting sentence per sub-claim.  
Although smaller in scale, WICE introduces more realistic cases where evidence is distributed across heterogeneous sources and claims are only partially supported.  

\paragraph{Benchmark construction.} 
\cref{tab:dataset_sizes} summarizes key statistics of our curated dataset.
By aggregating these datasets, our benchmark spans a spectrum of verification challenges: single-hop and multi-hop evidence, claims with partial evidence, and claims with multiple sufficient evidence sets, adding complexity to the ranking task. We selected 400 instances each from FEVER and HoVer to capture complementary characteristics of these large datasets. FEVER mainly provides single-hop evidence with multiple gold evidence sets, while HoVer offers multi-hop evidence with a single gold evidence set. From the smaller WICE dataset, we sampled 200 instances to maximize diversity, covering single- and multi-hop evidence, varying numbers of gold evidence sets, and partial evidence.  
To validate the reliability of the gold evidence, we manually inspected random samples across datasets and confirmed that the annotated sets are sufficient, non-contradictory, and exhaustive with respect to each claim. Full details are provided in \cref{sec:benchmark_construction}.

\begin{figure}[t]
\centering
\includegraphics[width=\columnwidth]{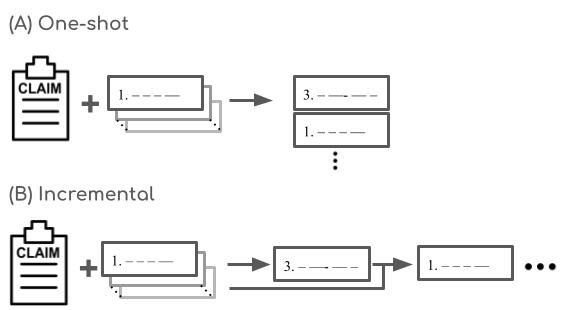}
\caption{Illustration of our proposed incremental ranking approach. (A) The naive one-shot algorithm produces a complete, global ranking of all evidence sentences in a single forward pass. (B) The incremental algorithm iteratively builds the ranking.}
\label{fig:methods}
\end{figure}

\section{Modeling}
\label{sec:modeling}

To explore the proposed Evidence Ranking task, we follow previous work (\cref{sec:background}) and consider four families of approaches that differ in how they assess the relationship between claims and candidate evidence: (i) embedding-based similarity, (ii) fine-tuned natural language inference (NLI) models, (iii) fine-tuned reasoning-based rerankers and (iv) large language models (LLMs).  
Each family can be instantiated in two algorithmic modes: \textit{one-shot}, where a full ranking is produced in a single pass, and \textit{incremental}, where sentences are ranked iteratively by conditioning on previously selected evidence (\cref{fig:methods}).  
This unified view allows us to compare lightweight baselines with more sophisticated methods under the same framework (See \cref{sec:appendix-methods} for full implementation details).

\subsection{Embedding-based Methods}
Embedding similarity provides a lightweight proxy for evidence relevance.  
Candidate sentences and the claim are encoded using a pretrained embedding model, and cosine similarity is used to rank sentences by closeness to the claim.
While efficient, this approach is local, evaluating each sentence independently and ignoring redundancy or complementarity across evidence.  
To address this, we also implement an incremental variant: after selecting the most similar sentence, subsequent selections are made by averaging each remaining candidate's embedding with those of the already-selected sentences and computing the cosine similarity to the claim embedding.

\textbf{Instantiation.} \texttt{BAAI/bge-large-en-v1.5} \cite{baai2023bge_large_en_v1_5} embedding model is used for its strong results in the Massive Text Embedding Benchmark \cite{muennighoff-etal-2023-mteb}.

\subsection{Fine-tuned NLI-based Methods}
Natural Language Inference (NLI) models are trained to identify entailment, contradiction, and neutral relationships, making them natural candidates for evidence ranking.
Following \citet{schuster-etal-2022-stretching}, we score each candidate sentence against the claim with a fine-tuned NLI model, obtaining separate rankings for \texttt{ENTAILS} and \texttt{CONTRADICTS}.
To select which ranking to use, we take the top-$k$ sentences from each (\(k=2\)) and form two short texts with \texttt{ENTAILS}-first and \texttt{CONTRADICTS}-first orderings. The NLI model scores both texts; we average label scores, select the higher-scoring label, and rank all sentences according to their score under the selected label.

Since NLI models are typically fine-tuned on single-sentence premises, there is no straightforward approach to create an incremental, global NLI-method that could improve over the local, one-shot approach. This limitation is further explored in \cref{sec:fine-tuned_nli_based_models_discussion}.

\textbf{Instantiation.} \texttt{NLI-DeBERTa-v3-large} entailment model is fine-tuned on multiple NLI datasets, and is used for its strong results on the SNLI and MNLI datasets \citep{reimers-2019-sentence-bert}.

\subsection{Fine-tuned Reasoning-Based Reranker Methods}
Fine-tuned reasoning-based rerankers generate a global ranking of candidate sentences given a claim, making them well-suited for evidence selection tasks that require prioritizing informative and non-redundant sentences. Unlike NLI-based methods, which score each sentence individually, rerankers directly output an ordered list of candidates.
We use the same reranker for both stages: first, an iterative filtering process is applied to identify the top 20 candidates, and then a one-shot global ranking is produced over these 20 sentences. Since fine-tuned rerankers are trained with a listwise objective, they rely on joint comparison of candidate sentences; consequently, there is no natural incremental variant, as ranking sentences one at a time removes the relative context required for accurate ordering.

\textbf{Instantiation.}  
We use \texttt{ReasonRank-7B} \cite{liu2025reasonrankempoweringpassageranking}, a fine-tuned model specialized for reasoning-based evidence ranking. The model outputs a global ranking of candidate sentences that favors informative and non-redundant evidence.

\subsection{LLM-based Methods}
LLMs demonstrate strong zero-shot reasoning abilities, particularly in NLI and fact verification tasks \cite{Zhang2023InterpretableULA, pisarevskaya2025zero}.  
Unlike previous local approaches, LLMs consider all candidate sentences jointly, producing a global ranking that prioritizes sufficiency and avoids redundancy.
In the incremental variant, the LLM iteratively selects the most informative sentence conditioned on those already chosen. 

\textbf{Instantiation.}  
For our main experiments, we use \texttt{GPT-4o} \cite{hurst2024gpt} and the open-source \texttt{Qwen3-235B} \cite{yang2025qwen3technicalreport}, both of which demonstrate strong capabilities in factual reasoning and evidence selection. We include both a closed commercial model and an open-source model to ensure reproducibility and wider accessibility.
To further study the effect of architecture, scale, and explicit reasoning capabilities, we also experiment with alternative LLMs and large reasoning models (LRMs) (\cref{sec:llms_lrms_analysis}).

\section{Results and Analyses}

\subsection{Main Results}
Our main results are summarized in \cref{tab:ranking-results}. We next analyze the key differences between the ranking methods and the insights revealed by our metrics.

\begin{table}[t]
\centering
\adjustbox{max width=\linewidth}{
    \begin{tabular}{lccc}
    \toprule
    \textbf{Ranking Method} & \makecell{\textbf{MRR $\uparrow$}} 
    & \makecell{\textbf{SR (\%) $\uparrow$}} \\
    \midrule
    One-Shot Similarity & 0.47\textsuperscript{\tiny ±0.01} & 32.4\textsuperscript{\tiny ±1.5} \\
    Incremental Similarity & 0.49\textsuperscript{\tiny ±0.01} & 33.5\textsuperscript{\tiny ±1.5} \\
    \midrule
    One-Shot Fine-tuned NLI & 0.31\textsuperscript{\tiny ±0.01} & 22.5\textsuperscript{\tiny ±1.3} \\
    \midrule
    One-Shot Fine-tuned Reranker & 0.67\textsuperscript{\tiny ±0.01} & 55.1\textsuperscript{\tiny ±1.5} \\
    \midrule
    \multicolumn{3}{l}{\textbf{GPT-4o}} \\
    One-Shot LLM & 0.69\textsuperscript{\tiny ±0.01} & 56.2\textsuperscript{\tiny ±1.6} \\
    Incremental LLM & \textbf{0.75}\textsuperscript{\tiny ±0.01} & \textbf{62.9}\textsuperscript{\tiny ±1.5} \\
    \multicolumn{3}{l}{\textbf{Qwen3-235B-A22B-Instruct}} \\
    One-Shot LLM & 0.71\textsuperscript{\tiny ±0.01} & 60.0\textsuperscript{\tiny ±1.5} \\
    Incremental LLM & \textbf{0.75}\textsuperscript{\tiny ±0.01} & 62.4\textsuperscript{\tiny ±1.5} \\
    \bottomrule
    \end{tabular}
}
\caption{Evaluation results across different ranking methods along with their standard error (SEM, see Appendix~\ref{sec:sem_computation} for details).} 
\label{tab:ranking-results}
\end{table}

\paragraph{LLM-based methods achieve the strongest overall performance.}
LLM-based approaches consistently outperform all other methods across all metrics. Incremental LLM ranking with GPT-4o achieves an MRR of 0.75 and a success rate of 62.9\%, while incremental Qwen3-235B achieves the same MRR and a slightly lower success rate of 62.4\%.
These results demonstrate LLMs can effectively capture both highly informative and complementary sentences, while avoiding redundancies.\footnote{See Appendix~\ref{sec:ndcg_results} for complementary NDCG analysis.}

\paragraph{Incremental strategies outperform non-incremental ones.}  
Leveraging previously selected sentences improves the ability to prioritize complementary evidence and avoid redundancy. For GPT-4o, incremental ranking improves MRR from 0.69 to 0.75 and raises the success rate from 56.2\% to 62.9\%. Qwen3-235B shows a similar pattern, with MRR increasing from 0.71 to 0.75 and success rate from 60.0\% to 62.4\%. Similarity-based methods exhibit the same trend, though with smaller gains: MRR increases from 0.47 to 0.49 and success rate from 32.4\% to 33.5\%, showing that even simple ranking approaches benefit from leveraging previously selected sentences.

\paragraph{One-shot fine-tuned reranker remains limited compared to incremental LLM ranking.}
Despite being a task-specific model fine-tuned for listwise ranking, the fine-tuned ReasonRank-7B reranker underperforms LLM-based methods. While its performance is competitive with one-shot general-purpose LLMs, it consistently falls short of the strongest LLM results in both MRR and success rate. This finding indicates that task-specific ranking fine-tuning alone is insufficient to match the evidence prioritization capabilities of large general-purpose LLMs in verification settings.

\paragraph{Similarity-based methods outperform fine-tuned NLI-based approaches.} 
The one-shot similarity-based method achieves 0.47 MRR and 32.4\% success rate, compared to 0.31 MRR and 22.5\% success rate for fine-tuned NLI ranking. In \cref{sec:fine-tuned_nli_based_models_discussion}, we explore this limitation and find that NLI models mostly fail in instances with multi-sentence evidence.

\paragraph{There remains headroom for improvement.}
\phantomsection
\label{par:headroom_improvement}

In \cref{tab:mrr_table} we report MRR segmented by the size of the minimal sufficient set. We find a significant decrease in performance as the requirement for more evidence sentences increases. When 3+ sentences are necessary for verification, MRR reaches a maximum of 0.52, compared to 0.94 for 1 sentence. 
In addition, as shown in Table~\ref{tab:ranking-results}, the success rate peaks at only 62.9\%, reflecting that even the best methods do not always rank all critical sentences early. These results highlight that while current approaches substantially reduce the reading burden compared to weaker baselines, achieving fully efficient coverage remains an open challenge.

\begin{figure*}[t]
\centering
\includegraphics[width=\textwidth]{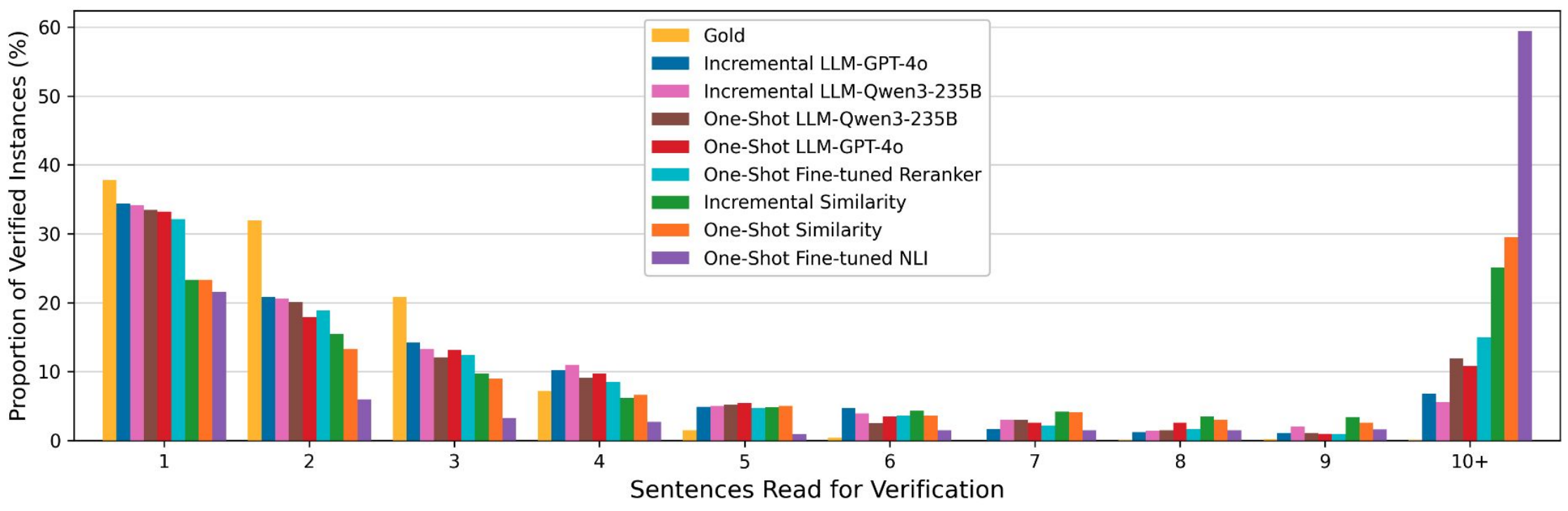}
\caption{Proportion of verified instances as a function of the number of sentences read (non-cumulative).}
\label{fig:bar_graph_ranking_methods}
\end{figure*}

\begin{table}[t]
\centering
\adjustbox{max width=\linewidth}{
    \begin{tabular}{lccc}
    \toprule
    \textbf{Method} & \textbf{1} & \textbf{2} & \textbf{3+} \\
    \midrule
    One-Shot Similarity & 0.74\textsuperscript{\tiny ±0.01} & 0.39\textsuperscript{\tiny ±0.01} & 0.19\textsuperscript{\tiny ±0.01} \\
    Incremental Similarity & 0.74\textsuperscript{\tiny ±0.01} & 0.44\textsuperscript{\tiny ±0.02} & 0.18\textsuperscript{\tiny ±0.01} \\
    \midrule
    One-Shot Fine-tuned NLI & 0.67\textsuperscript{\tiny ±0.02} & 0.12\textsuperscript{\tiny ±0.01} & 0.05\textsuperscript{\tiny ±0.00} \\
    \midrule
    One-Shot Fine-tuned Reranker & 0.90\textsuperscript{\tiny ±0.01} & 0.64\textsuperscript{\tiny ±0.02} & 0.39\textsuperscript{\tiny ±0.02} \\
    \midrule
    \multicolumn{3}{l}{\textbf{GPT-4o}} \\
    One-Shot LLM & 0.92\textsuperscript{\tiny ±0.01} & 0.64\textsuperscript{\tiny ±0.02} & 0.44\textsuperscript{\tiny ±0.02} \\
    Incremental LLM & \textbf{0.94}\textsuperscript{\tiny ±0.00} & \textbf{0.73}\textsuperscript{\tiny ±0.01} & 0.51\textsuperscript{\tiny ±0.02} \\
    \multicolumn{3}{l}{\textbf{Qwen3-235B-A22B-Instruct}} \\
    One-Shot LLM & 0.93\textsuperscript{\tiny ±0.01} & 0.67\textsuperscript{\tiny ±0.02} & 0.47\textsuperscript{\tiny ±0.02} \\
    Incremental LLM & \textbf{0.94}\textsuperscript{\tiny ±0.00} & 0.71\textsuperscript{\tiny ±0.01} & \textbf{0.52}\textsuperscript{\tiny ±0.02} \\
    \bottomrule
    \end{tabular}
}
\caption{MRR of each ranking method for instances grouped by the size of their optimal gold evidence set, along with their standard error (SEM, see Appendix~\ref{sec:sem_computation} for details).}
\label{tab:mrr_table}
\end{table}

\subsection{Analysis of the User Reading Effort}\label{sec:segmentation_analysis}

To better understand the user experience when interacting with evidence ranking methods, we report the proportion of instances verified as a function of the number of sentences that the user has to read (\cref{fig:bar_graph_ranking_methods}). In \cref{sec:user_study}, we describe a user study with actual users, which supports the results of this analysis.

Although most claims can be verified with three sentences or fewer, and the optimal evidence requires at most six, we find that all methods occasionally require reading ten or more sentences. This means that users often review substantially more sentences than necessary and may withdraw before seeing all relevant evidence. We include in the Appendix an accumulative variant of this figure (\cref{fig:recall_sentences_count_methods}), which provides more insights on model behavior.

\subsection{Exploration of different LLMs and LRMs}\label{sec:llms_lrms_analysis}

Given that the incremental LLM-based approach performs better than others, we set out to explore the differences between LLM architectures, model sizes, and LLMs versus large reasoning models (LRMs), providing a deeper understanding of how these model characteristics influence ranking performance.
We evaluated four additional non-reasoning LLMs \citep{openai20244o, llama2024llama3_3, llama2025llama4, yang2025qwen3technicalreport} and three LRMs \cite{openai2025o3o4mini, openai2025o3mini, yang2025qwen3technicalreport}. Due to the high cost of closed-source models, we used only half of the dataset described in \cref{sec:datasets}.

The results are reported in Table~\ref{tab:llm_lrm_results}.  \texttt{o3} and  \texttt{o3-mini} achieve the strongest results across metrics, 
indicating that the reasoning-optimized  \texttt{o3} line provides tangible improvements over  \texttt{4o} and  \texttt{4o-mini}, with the more compact  \texttt{o3-mini} variant maintaining performance close to the full  \texttt{o3} model.
The open-source  \texttt{LLaMA} and  \texttt{Qwen3} LLMs perform competitively with \textsc{4o}.
By contrast, the  \texttt{Qwen3-235B-Thinking} model exhibits a notable drop in performance.
In our manual inspection, we observed that it frequently provides detailed per-evidence reasoning, but it fails to output a selection for the next sentence. We further discuss this in \cref{sec:qwen3_output_example}.

\begin{table}[t]
\centering
\small
\adjustbox{max width=\linewidth}{
    \begin{tabular}{lccc}
    \toprule
    & \makecell{\textbf{MRR} $\uparrow$}
    & \makecell{\textbf{SR (\%)} $\uparrow$} \\
    \midrule
    \multicolumn{3}{l}{\textbf{LLMs}} \\
    4o & 0.83\textsuperscript{\tiny ±0.01} & 73.0\textsuperscript{\tiny ±2.0} \\
    4o-mini & 0.79\textsuperscript{\tiny ±0.01} & 68.0\textsuperscript{\tiny ±2.0}  \\
    LLaMA-3.3-70B & 0.83\textsuperscript{\tiny ±0.01} & 74.0\textsuperscript{\tiny ±2.0} \\
    LLaMA-4-Maverick-17B & 0.80\textsuperscript{\tiny ±0.01} & 70.0\textsuperscript{\tiny ±2.0} \\
    Qwen3-235B-A22B-Instruct & 0.83\textsuperscript{\tiny ±0.01} & 73.8\textsuperscript{\tiny ±2.0} \\
    \midrule
    \multicolumn{3}{l}{\textbf{LRMs (Reasoning Models)}} \\
    o3 & \textbf{0.88}\textsuperscript{\tiny ±0.01} & \textbf{79.4}\textsuperscript{\tiny ±1.8} \\
    o3-mini & 0.87\textsuperscript{\tiny ±0.01} & 78.4\textsuperscript{\tiny ±1.8} \\
    Qwen3-235B-A22B-Thinking & 0.63\textsuperscript{\tiny ±0.01} & 52.6\textsuperscript{\tiny ±2.2} \\
    \bottomrule
    \end{tabular}
}
\caption{Performance of different LLMs and LRMs using the incremental LLM approach.}
\label{tab:llm_lrm_results}
\end{table}

\subsection{Fine-tuned NLI-based Models}
\label{sec:fine-tuned_nli_based_models_discussion}

\cref{tab:mrr_table} reveals a key limitation of fine-tuned NLI models: they struggle to rank multi-sentence evidence, even when compared to the simpler similarity-based approach. We hypothesize that this stems from a fundamental limitation of these fine-tuned NLI models: they frequently struggle to assign sufficiently high scores to partially supporting sentences when compared to neutral ones. Specifically, if a sentence contains crucial verification details but is predominantly neutral in its overall sentiment, the fine-tuned NLI model may assign it a relatively weak score for either support or refute. This makes it difficult for such sentences to be effectively surfaced in ranking. In contrast, similarity-based ranking, by leveraging semantic or lexical closeness, often succeeds in grouping together more sentences that collectively support or refute the claim. Furthermore, our experimentation with an incremental version of the NLI approach also showed it to underperform the one-shot method, with more details provided in \cref{sec:incremental_nli_analysis}.

\subsection{Qualitative Analysis}
\label{sec:qalitive_analysis_discussion}

To further corroborate our main takeaways, we conducted a manual analysis of a representative example, tracing the ranking order for each method until a sufficient evidence set was fully covered (though not necessarily the optimal one). This example highlights key differences between methods, as it contains multiple gold evidence sets and an optimal gold evidence set requiring more than one sentence, properties that make it well suited to reveal variations in efficiency and ranking behavior. The incremental LLM-based method ranked key evidence earlier while largely avoiding both redundant and irrelevant sentences. The one-shot LLM-based approach, as well as the fine-tuned reranker, behaved similarly, occasionally including irrelevant sentences. Similarity-based methods tended to surface more irrelevant sentences, while fine-tuned NLI-based methods were the least efficient, often requiring review of many unrelated sentences. The distinction between redundant and irrelevant sentences is important: redundant sentences provide relevant information but are unnecessary if similar evidence has already been presented, whereas irrelevant sentences do not contribute to verifying the claim at all. A detailed analysis of this example, including the full ranking order, is provided in \cref{sec:detailed_qualitative_analysis}.

\section{User Study}\label{sec:user_study}

We conducted a controlled user study to assess whether incremental evidence ranking reduces user effort while maintaining verification accuracy. The study focuses on two central research questions. First, we evaluate whether evidence ranking improves efficiency without harming accuracy. Second, we examine whether users can successfully identify the stopping point where enough evidence was presented to make a verification decision. 
Together, these evaluate both the practical benefits of incremental evidence ranking and the validity of the ``minimal sufficiency'' assumption that underlies our task formulation.

\subsection{Setup}

Five fluent English speakers with a computer-science background participated in the study. We used 100 claims from our curated dataset and evaluated two systems:
\begin{itemize}
    \item \textbf{Evidence ranking:} We utilized the incremental LLM evidence ranking with GPT-4o.
    \item \textbf{LLM evidence selection:} We utilized the LLM evidence selection method used in \citep{hirsch-etal-2025-laquer}, adapted to the sentence-level.
\end{itemize}

For evidence ranking, our interface allows users to consume ranked evidence incrementally and stop when the information encountered feels sufficient. At this point, users must make a verification decision.
For evidence selection, the interface presents all selected evidence, and the user has to make a verification decision based only on that evidence. To emphasize, in the evidence selection variant, the users have no control on the number of sentences they got.

Further details on the study design, trial assignment, and interface are provided in Appendix~\ref{app:user_study_detailed}.

\subsection{Results}

To assess how users interact with different evidence presentation methods, we analyzed verification accuracy and reading effort across the two conditions. If participants chose \emph{can't decide} before reading all available evidence sentences, we interpret this as the user “giving up” on verifying the claim, effectively concluding that it is neither supported nor refuted.

Table~\ref{tab:user_study_results} summarizes the main findings. Participants achieved higher verification success with evidence ranking (94\%) compared to evidence selection (74\%). Incorrect label choices were rare in both settings (at most 3 cases per condition), while evidence selection produced substantially more \emph{can't decide} outcomes (23\% vs.\ 5\%), suggesting that its evidence sets were often insufficient for users to reach a definitive verdict.

The evidence ranking approach also reduced the reading effort: participants read an average of 2.48 sentences before making a decision, compared to 3.85 with evidence selection. To contextualize this behavior, we compared the number of sentences that users read to the sizes of the minimal gold-sufficient evidence sets. Based on gold annotations, users are required to read on average only 1.88 sentences for complete verification, yet read 2.48, which represents an average surplus of 0.60 sentences. This suggests that users typically read slightly beyond the minimal evidence to increase confidence, while still benefiting from substantial efficiency gains.

We further analyzed how often participants read more evidence than was necessary. The examples over-read rate was 10\% for evidence ranking, compared to 35\% for evidence selection. This demonstrates that incremental ranking not only improves efficiency, supporting our first research question, but also helps users identify the appropriate stopping point for verification, addressing our second question.

Overall, the results show that incremental evidence ranking enhances verification accuracy and reduces reading effort, supporting our first research question on efficiency and our second question on users' ability to identify sufficient evidence. These findings further validate the motivation for designing evidence ranking methods.

\begin{table}[t]
\centering
\large
\adjustbox{max width=\linewidth}{
    \begin{tabular}{lcc}
    \toprule
    \textbf{Metric} & \textbf{Evidence Ranking} & \textbf{Evidence Selection} \\
    \midrule
    Success rate & 94\%  & 74\% \\
    Wrong decision rate & 1\%  & 3\% \\
    Undecided rate & 5\%  & 23\% \\
    \hline
    Average sentences read & 2.48 & 3.85 \\
    Examples over-read & 10\% & 35\% \\
    \bottomrule
    \end{tabular}
}
\caption{User study results comparing LLM incremental evidence ranking with LLM evidence selection. The gold average number of sentences is 1.88. Over-read refers to instances where the user read more sentences than necessary.}
\label{tab:user_study_results}
\end{table}

\section{Conclusions}
In this work, we discussed the user's perspective in both human fact-checking and LLM attribution scenarios, and repurposed the evidence selection task as an evidence ranking task.
We proposed novel evaluation metrics and experimented with both existing and novel modeling approaches.
Our key results show that LRMs work best and that ranking incrementally can further improve results.
LRMs are notoriously costly, and future work could focus on methods fine-tuned for the evidence ranking task.
In addition, future work could experiment with datasets where evidence can simultaneously include both supporting and contradictory statements \citep{nachshoni2025consensus}. Overall, our results underscore the feasibility and significant potential of this task to deliver substantial benefits to users. \todo{another future work could be hybrid work}

\FloatBarrier
\clearpage
\newpage
\section*{Limitations}

Our work focuses on standard verification datasets with human-annotated evidence, which provide a well-controlled environment for evaluating evidence ranking. However, these datasets do not supply any internal ordering of evidence sentences, meaning that coherence-based ordering cannot be evaluated within our current setup. Developing and assessing methods for producing coherent evidence orderings is therefore an important direction for future work. The datasets also primarily contain short, focused claims, and extending our approach to longer or more complex claims is a natural next step for future research.
We additionally conducted a user study to examine how our ranking methods support users in practice. While informative, the study represents an initial exploration, involving a small participant group and a curated subset of examples. Future studies with larger, more diverse populations would allow for a more comprehensive understanding of the practical benefits of the proposed approach.

\section*{Ethical Considerations}

Our task simulates the process a user goes through when verifying claims, reading evidence until sufficient information is found to support or refute a claim. Errors in the evidence ranking could mislead users into assuming a stronger or weaker connection between the claim and the retrieved evidence.

AI-assisted writing tools were used to improve clarity and coherence, and all content was reviewed and edited by the authors for accuracy.

The qualitative analysis and user study reported in this work involved manual annotation by five undergraduate Computer Science students. The participants were recruited internally from the authors' institution. Before the study, all participants were informed about the research goals and the intended use of the collected data. Participation was strictly voluntary; no financial compensation was provided, and the study was conducted in accordance with the institution's ethical guidelines. All data was collected and analyzed anonymously.

\section*{Acknowledgments}
We would like to thank our reviewers for their
constructive suggestions and comments. This work was supported in part by the Israel Science Foundation (grants no. 2827/21 and 3182/25).
SB is funded by the German Federal Ministry of Research, Technology and Space (BMFTR) under the promotional reference 01ZZ2314H (GeMTeX).

\bibliography{references}

\clearpage
\newpage
\appendix

\lstdefinestyle{promptStyle}
{
    basicstyle={\footnotesize\ttfamily},%
    numbers=none,
    numberstyle=\footnotesize,
    xrightmargin=1.5em,
    showstringspaces=false,
      showspaces=false,
        showtabs=false,
    tabsize=2,
    breaklines=false,
        flexiblecolumns=true,
        escapeinside={<@}{@>},
          breakatwhitespace=true
}

\newtcblisting{mylisting}[1]{
  enhanced,
  listing only,
  boxrule=0.8pt,
  sharp corners=downhill,
  top=0mm,
  bottom=0mm,
  left=2mm,
  right=0mm,
  boxsep=0mm,
  colframe=black,
  colback=white,
  listing options={
    style=#1
  }
}

\definecolor{instructionsColor}{rgb}{0.1, 0.5, 0.1}
\definecolor{darkyellow}{rgb}{0.7,0.6,0.0}

\section{Task Definition}
\label{app:task-definition}
Following is a formal \textbf{Evidence Ranking} task definition. 

Given a claim $c$ and a set of candidate evidence sentences 
$S = \{s_1, \dots, s_n\}$, the goal is to produce a ranked ordering $P = \langle s_{p_1}, \dots, s_{p_n} \rangle$ of $S$.  
At any decision point $i$, the accumulated prefix set 
$P_i = \{s_{p_1}, \dots, s_{p_i}\}$, may be sufficient.
We say that a prefix $P_i$ is \emph{sufficient} for claim $c$ if the there exists a subset $E \subseteq P_i$ such that $E$ confidently verify or refute the claim $c$: 
\begin{equation}
\begin{split}
\mathrm{Suf}(P_i, c) \quad \equiv & \quad \exists E \subseteq P_i \quad \text{such that} \\ & (E \implies c)  \lor (E \implies \lnot c)
\end{split}
\end{equation}

The \emph{Minimal Sufficient Rank} (MSR) is defined as follows:
\begin{equation}
\mathrm{MSR}(P, c) \;=\; 
\min \Bigl\{\, i \;\big|\;  \mathrm{Suf}(P_i, c) \,\Bigr\}
\end{equation}

We note that this definition holds only for cases in which there is a sufficient set of evidence. As this is there case in our dataset the MSR is well defined.

The objective of the Evidence Ranking task is to produce an ordering that minimizes $\mathrm{MSR}$.

\subsection{Mean Reciprocal Rank (MRR).}
\label{app:MRR-definition}
For a given claim $c$ and a ranked ordering $P$ of evidence sentences $S$, we calculate the corresponding Minimal Sufficient Rank and compare it to the Ideal Minimal Sufficient Rank:
\[
\mathrm{rank} \;=\; \mathrm{MSR}(P, c) - \mathrm{IMSR}(S, c) + 1
\]
where 
\[
\mathrm{IMSR}(S, c) \;=\; 
\min_{P \in \Pi(S)} \mathrm{MSR}(P, c),
\]
Dataset-level performance is then measured by the Mean Reciprocal Rank (MRR), defined as the average reciprocal rank ($\frac{1}{rank}$) across all claims.

\subsection{Success Rate}
\label{app:SR-definition}
Formally, given the Ideal Minimal Sufficient Rank (IMSR) of a candidate set $S$ for claim $c$, $SR$ is defined as the sufficiency at $IMSR$:
\[
\mathrm{SR}(P, c) \;=\; \mathrm{Suf}\!\left(P_{\mathrm{IMSR}(S, c)}, c\right).
\]

\subsection{Normalized Discounted Cumulative Gain (NDCG)}
\label{app:NDCG-definition}

We adapt Normalized Discounted Cumulative Gain (NDCG) \cite{10.1145/582415.582418} to evaluate the quality of evidence rankings beyond identifying only a minimal sufficient prefix. While sufficiency-based metrics focus on how quickly verification becomes possible, NDCG rewards rankings that place all relevant evidence sentences early, capturing the overall usefulness of the ranking.

For a given claim $c$ and a ranked list of evidence sentences $P = (s_{p_1}, \ldots, s_{p_n})$, relevance is determined in two steps. First, we restrict attention to the minimally sufficient prefix $P_{msr}$, where $msr = MSR(P, c)$ is the smallest rank at which the accumulated evidence is sufficient to verify or refute the claim. Second, within this prefix, we identify the minimal sufficient gold evidence set $G$, defined as:
\[
G \;=\; 
\arg\min_{E \in \Pi(P_{msr})} 
\Bigl\{\, |E| \;\big|\; \mathrm{Suf}(E, c) \,\Bigr\},
\]
where $\Pi(P_{msr})$ denotes all subsets of $P_{msr}$. The subset $G$ contains exactly the sentences required to justify the claim.

Only sentences in $G$ are treated as relevant; all other sentences are considered non-relevant. Given this relevance assignment, we compute the Discounted Cumulative Gain (DCG) over the prefix $P_{msr}$:
\[
\text{DCG} = \sum_{i=1}^{msr} \frac{rel_i}{\log_2(i+1)},
\]
where $rel_i = 1$ if $s_{p_i} \in G$ and $0$ otherwise. The final NDCG score is obtained by normalizing DCG by the DCG of the ideal ordering of $G$:
\[
\text{NDCG} = \frac{\text{DCG}}{\text{IDCG}}.
\]

This normalization bounds scores between 0 and 1, with higher values indicating that the relevant evidence sentences are ranked closer to the top.

\section{Benchmark Construction} 
\label{sec:benchmark_construction}
To create a robust benchmark for evaluating evidence ranking methods, we carefully selected instances from FEVER \citep{thorne-etal-2018-fever}, HoVer \citep{jiang-etal-2020-hover}, and WICE \citep{kamoi-etal-2023-wice} to ensure coverage of diverse verification scenarios.  FEVER and WiCE are under the Creative Commons Attribution-ShareAlike license, and HoVer under the CC BY-SA 4.0 license.

First, we considered the amount of evidence required for verification. Some claims can be verified using a single sentence, allowing for simple relevance assessment, while others span multiple documents and require multiple sentences to be considered together, testing the model's ability to account for redundancy and supplementary information. 

Second, we accounted for instances with multiple sets of sufficient evidence versus those with only a single sufficient set. This enables evaluation not only of whether a model identifies relevant evidence, but also whether it consistently selects sentences from the same set that collectively support or refute the claim. 

Third, we ensured representation of both supporting and refuting evidence cases, reflecting the full range of verification scenarios. 

Accordingly, instances were sampled from each dataset while enforcing constraints to guarantee diversity across these characteristics. Approximately 60\% of the selected instances contain a single sufficient evidence set, with the remainder containing 2-4 sets. The optimal number of sentences in a sufficient set is distributed as follows: roughly one third of instances have size 1, one third size 2, 20\% size 3, and the remaining instances contain 4-9 sentences. Overall, the benchmark includes evidence spanning from single sentences to larger sets of up to approximately 60 sentences. 

This selection strategy ensures that the benchmark evaluates model performance across a wide spectrum of verification challenges, including single- versus multi-sentence evidence, multiple evidence sets, and both supporting and refuting cases, thereby providing a comprehensive assessment of evidence ranking methods.

\section{Ranking Methods Implementation}
\label{sec:appendix-methods}

In this section, we provide a detailed description of the ranking methods introduced in Section 4.

\begin{figure*}[p]
\begin{mylisting}{promptStyle}

<@\textcolor{instructionsColor}{You are given:}@>
<@\textcolor{instructionsColor}{1. A factual statement.}@>
<@\textcolor{instructionsColor}{2. A list of exactly 9 numbered sentences extracted from source documents. Sentence IDs start at 1.}@>

<@\textcolor{instructionsColor}{Rank all 9 sentences by how directly and clearly they relate to the factual statement. A sentence relates to the statement if it clearly provides evidence about it, without requiring inference. Rank from the most directly informative sentence to the least, and include all sentences.}@>


<@\textcolor{instructionsColor}{\#\#\# OUTPUT FORMAT RULES \#\#\#}@>
<@\textcolor{instructionsColor}{- Output MUST be a JSON object with sentence IDs as keys and sentence texts as values.}@>
<@\textcolor{instructionsColor}{- The order of the key-value pairs in the JSON represents the ranking: the first pair is the most relevant sentence, the second is next, and so on.}@>
<@\textcolor{instructionsColor}{- Example: \{\{"1": "Sentence text 1", "5": "Sentence text 5", "3":
"Sentence text 3"\}\} means sentence 1 is ranked first, sentence 5 second, and sentence 3 third.}@>
<@\textcolor{instructionsColor}{- Do NOT output explanations, reasoning, or any extra text.}@>
<@\textcolor{instructionsColor}{- Your ENTIRE response must exactly match the required format.}@>

<@\textcolor{black}{Statement: Bill Lewis was the head coach of the Georgia Tech football team that finished the regular season with the same record as the 1981 team.}@>

<@\textcolor{black}{Sentences:}@>
<@\textcolor{black}{1. The 1981 Georgia Tech Yellow Jackets football team represented the Georgia Institute of Technology during the 1981 NCAA Division I-A football season.}@>
<@\textcolor{black}{2. The Yellow Jackets were led by second-year head coach Bill Curry, and played their home games at Grant Field in Atlanta, Georgia.}@>
<@\textcolor{black}{3. Georgia Tech produced abysmal results for the second consecutive year under Curry, finishing with a record of 1–10, their worst season in terms of winning percentage in school history (it would later be matched by another 1-10 season in 1994).}@>
<@\textcolor{black}{4. The Yellow Jackets were lead by head coach Bill Lewis through eight games, being fired after going 1-7.}@>
<@\textcolor{black}{...}@>

<@\textcolor{red}{Output:}@>
<@\textcolor{red}{
\{"1": "The 1981 Georgia Tech Yellow Jackets football team
represented the Georgia Institute of Technology during the 1981
NCAA Division I-A football season.",}@>
<@\textcolor{red}{"4": "The Yellow Jackets were lead by head coach Bill Lewis through
eight games, being fired after going 1-7.",}@>
<@\textcolor{red}{...}@>
<@\textcolor{red}{\}}@>


\end{mylisting}
\caption{Example prompt for LLM evidence ranking. The instructions are depicted in green, input to the model in black, and model's output in red.}
\label{fig:llm_ranking_prompt}
\end{figure*}

\subsection{LLM}
The prompt is shown in Figure~\ref{fig:llm_ranking_prompt}. We use GPT-4o and manually optimized the prompt instructions through iterative testing on development set examples that were not used in evaluation. For each model output, we ensured that all sentences were returned; if not, the model was prompted again, up to a maximum of five attempts. In cases where not all sentences were ranked after five attempts, the remaining unranked sentences were placed at the end of the ranking in their original reading order.

\begin{figure*}[p]
\begin{mylisting}{promptStyle}

<@\textcolor{instructionsColor}{You are given:}@>
<@\textcolor{instructionsColor}{1. A factual statement.}@>
<@\textcolor{instructionsColor}{2. A list of numbered sentences extracted from source documents.}@>
<@\textcolor{instructionsColor}{Identify exactly one sentence number from the list that directly and explicitly relates to the factual statement. A sentence relates to the statement if it clearly provides evidence about it, without requiring inference. Select the single most relevant sentence from the list, choosing the one that provides the clearest evidence.}@>

<@\textcolor{instructionsColor}{\#\#\# OUTPUT FORMAT RULES \#\#\#}@>
<@\textcolor{instructionsColor}{- Output MUST consist of ONLY one sentence number in square brackets.}@>
<@\textcolor{instructionsColor}{- Example: [2]}@>
<@\textcolor{instructionsColor}{- Do NOT output explanations, reasoning, or any extra text.}@>
<@\textcolor{instructionsColor}{- Your ENTIRE response must exactly match the required format.}@>

<@\textcolor{black}{Statement: Bill Lewis was the head coach of the Georgia Tech football team that finished the regular season with the same record as the 1981 team.}@>

<@\textcolor{black}{Sentences:}@>
<@\textcolor{black}{1. The 1981 Georgia Tech Yellow Jackets football team represented the Georgia Institute of Technology during the 1981 NCAA Division I-A football season.}@>
<@\textcolor{black}{2. The Yellow Jackets were led by second-year head coach Bill Curry, and played their home games at Grant Field in Atlanta, Georgia.}@>
<@\textcolor{black}{3. Georgia Tech produced abysmal results for the second consecutive year under Curry, finishing with a record of 1–10, their worst season in terms of winning percentage in school history (it would later be matched by another 1-10 season in 1994).}@>
<@\textcolor{black}{4. The Yellow Jackets were lead by head coach Bill Lewis through eight games, being fired after going 1-7.}@>
<@\textcolor{black}{...}@>

<@\textcolor{red}{Output:}@>
<@\textcolor{red}{[1]}@>


\end{mylisting}
\caption{Example prompt for the first step in the incremental LLM evidence ranking, in which we don't have any selected sentences yet. The instructions are depicted in green, input to the model in black, and model's output in red. This prompt is intended for selecting the first sentence freely, without restrictions, i.e., it does not depend on any previously chosen sentences.}
\label{fig:incremental_llm_prompt_first}
\end{figure*}

\begin{figure*}[p]
\begin{mylisting}{promptStyle}

<@\textcolor{instructionsColor}{You are given:}@>
<@\textcolor{instructionsColor}{1. A factual statement.}@>
<@\textcolor{instructionsColor}{2. A list of numbered sentences extracted from source documents.}@>
<@\textcolor{instructionsColor}{3. A list of sentences labeled "used" (given as raw text, not sentence numbers) that have already been selected as relevant evidence.}@>
<@\textcolor{instructionsColor}{Identify exactly one additional sentence number from the numbered list that, when combined with the "used" sentences, directly and explicitly relates to the factual statement. Do NOT re-select any sentences whose text appears in the "used" sentences list.}@>
<@\textcolor{instructionsColor}{Select the single most relevant new sentence from the list, choosing the one that provides the clearest evidence.}@>
<@\textcolor{instructionsColor}{\#\#\# OUTPUT FORMAT RULES \#\#\#}@>
<@\textcolor{instructionsColor}{- Output MUST consist of ONLY one newly selected sentence number in square brackets.}@>
<@\textcolor{instructionsColor}{- Example: [2]}@>
<@\textcolor{instructionsColor}{- Do NOT output explanations, reasoning, or any extra text.}@>
<@\textcolor{instructionsColor}{- Your ENTIRE response must exactly match the required format.}@>

<@\textcolor{black}{Statement: Bill Lewis was the head coach of the Georgia Tech football team that finished the regular season with the same record as the 1981 team.}@>

<@\textcolor{black}{Sentences:}@>
<@\textcolor{black}{1. The 1981 Georgia Tech Yellow Jackets football team represented the Georgia Institute of Technology during the 1981 NCAA Division I-A football season.}@>
<@\textcolor{black}{2. The Yellow Jackets were led by second-year head coach Bill Curry, and played their home games at Grant Field in Atlanta, Georgia.}@>
<@\textcolor{black}{3. Georgia Tech produced abysmal results for the second consecutive year under Curry, finishing with a record of 1–10, their worst season in terms of winning percentage in school history (it would later be matched by another 1-10 season in 1994).}@>
<@\textcolor{black}{4. The Yellow Jackets were lead by head coach Bill Lewis through eight games, being fired after going 1-7.}@>
<@\textcolor{black}{...}@>

<@\textcolor{red}{Output:}@>
<@\textcolor{red}{[3]}@>

\end{mylisting}
\caption{Example prompt for incremental LLM evidence ranking. The instructions are depicted in green, input to the model in black, and model's output in red. In this prompt, the selection depends on the sentences that have already been chosen and ranked in previous iterations.}
\label{fig:incremental_llm_prompt_second}
\end{figure*}

\begin{figure*}[p]
\begin{mylisting}{promptStyle}

<@\textcolor{instructionsColor}{You are RankLLM, an intelligent assistant that can rank sentences based on their relevance to a factual statement.}@>
<@\textcolor{instructionsColor}{Given a factual statement and a sentence list, first think about the reasoning process in your mind, and then provide the reranked sentence list.}@>
<@\textcolor{instructionsColor}{Enclose the reasoning and answer within <think> and <answer> tags, respectively, i.e., <think> reasoning here </think> <answer> answer here </answer>.}@>
<@\textcolor{instructionsColor}{You will be provided 9 sentences, each indicated by a numerical identifier.}@>
<@\textcolor{instructionsColor}{Rank the sentences based on their relevance to the factual statement.}@>

<@\textcolor{black}{Sentences:}@>
<@\textcolor{black}{1. The 1981 Georgia Tech Yellow Jackets football team represented the Georgia Institute of Technology during the 1981 NCAA Division I-A football season.}@>
<@\textcolor{black}{2. The Yellow Jackets were led by second-year head coach Bill Curry, and played their home games at Grant Field in Atlanta, Georgia.}@>
<@\textcolor{black}{3. Georgia Tech produced abysmal results for the second consecutive year under Curry, finishing with a record of 1–10, their worst season in terms of winning percentage in school history (it would later be matched by another 1-10 season in 1994).}@>
<@\textcolor{black}{4. The Yellow Jackets were lead by head coach Bill Lewis through eight games, being fired after going 1-7.}@>
<@\textcolor{black}{...}@>

<@\textcolor{black}{Factual Statement: Bill Lewis was the head coach of the Georgia Tech football team that finished the regular season with the same record as the 1981 team.}@>

<@\textcolor{instructionsColor}{All the sentences should be included and listed using identifiers, in descending order of relevance.}@>
<@\textcolor{instructionsColor}{The format of the answer should be [] > [], e.g., [2] > [1].}@>

<@\textcolor{red}{Output:}@>
<@\textcolor{red}{<think> reasoning about which sentences best support or refute the statement </think>}@>
<@\textcolor{red}{<answer> [1] > [4] > [2] > [3] > ... </answer>}@>

\end{mylisting}
\caption{Example prompt for fine-tuned reasoning-based reranker (ReasonRank-7B). The instructions are depicted in green, input sentences in black, and model output in red. The reranker processes up to 20 sentences per batch, with reasoning enclosed in <think> tags and final ranking in <answer> tags.}
\label{fig:reranker_prompt}
\end{figure*}

\subsection{Incremental LLM}
The incremental LLM ranking uses two prompts. The first prompt, shown in Figure~\ref{fig:incremental_llm_prompt_first}, identifies the sentence most relevant to supporting or refuting the claim. Subsequent prompts, shown in Figure~\ref{fig:incremental_llm_prompt_second}, are applied iteratively: at each step, the model selects the next sentence that, together with the previously selected sentences, directly and explicitly relates to the claim. We applied this procedure to all LLMs described in the study, with prompts manually optimized on development set examples not used in evaluation. The model was always forced to return a sentence; if no sentence was returned after five attempts, any remaining unselected sentences were appended in their original reading order.

\subsection{Fine-tuned Reasoning-Based Reranker}
\label{app:reranker_appendix}
The prompt is shown in Figure~\ref{fig:reranker_prompt}. We use ReasonRank-7B and follow the prompt pattern from \cite{liu2025reasonrankempoweringpassageranking}, manually optimizing instructions through iterative testing on development set examples not used in evaluation. To respect the model's fine-tuning constraint, which optimizes ranking for a maximum of 20 items, we implement a Tournament-style Survival Ranking: we maintain a "top-tier" window of 20 sentences and iteratively replace the weakest candidate with the next unseen sentence to identify the best 20 sentences. The final output is a single one-shot ranking of these top 20 sentences; the iterative process is only used to select them, not to produce an incremental ranking. This ensures the model operates within its optimal range while producing a meaningful global ranking.

\subsection{Similarity}

\begin{algorithm}[H]
\small
\caption{Similarity-based Ranking}\label{alg:similarity}
\begin{algorithmic}
\State $\textit{scores} \gets \{\}$ 
\State $\textit{claim\_emb} \gets \text{Embed}(\textit{claim})$
\State $\textit{evidence\_embs} \gets \text{EmbedAll}(\textit{evidence\_list})$
\For{$\textit{emb} \in \textit{evidence\_embs}$}
    \State $s \gets \text{CosSim}(\textit{claim\_emb}, \textit{emb})$
    \State $\textit{scores[emb]} \gets s$
\EndFor
\State $\textit{ranked\_evidence} \gets \text{SortDescending}(\textit{scores})$
\State \textbf{return} $\textit{ranked\_evidence}$
\end{algorithmic}
\end{algorithm}

\paragraph{Explanation.} 
The claim and candidate evidence sentences are first encoded into dense vector representations using the \texttt{BAAI/bge-large-en-v1} embedding model. 
Cosine similarity is then computed between the claim embedding and each evidence embedding, producing a similarity score for every candidate. 
These scores are stored in a dictionary mapping each evidence sentence to its similarity value. 
Finally, the evidence is ranked in descending order of similarity, ensuring that sentences most relevant to the claim appear first.

\subsection{Incremental Similarity-based Ranking}

\begin{algorithm}[H]
\small
\caption{Incremental Similarity-based Ranking}\label{alg:incremental-similarity}
\begin{algorithmic}
\State $\textit{ranked\_evidence} \gets []$
\State $\textit{claim\_emb} \gets \text{Embed}(\textit{claim})$
\State $\textit{evidence\_embs} \gets \text{EmbedAll}(\textit{evidence\_list})$
\State $\textit{selected\_indices} \gets \{\}$

\For{$i = 1$ to length of $\textit{evidence\_list}$}
    \State $\textit{best\_idx} \gets$ None
    \State $\textit{best\_score} \gets -1$
    \For{each $idx, emb$ in $\textit{evidence\_embs}$}
        \If{$idx \in \textit{selected\_indices}$} \textbf{continue} \EndIf
        \State $\textit{avg\_emb} \gets$ avg($\textit{ranked\_evidence\_embs} + emb$)
        \State $\textit{sim\_score} \gets \text{CosSim}(\textit{claim\_emb}, \textit{avg\_emb})$
        \If{$\textit{sim\_score} > \textit{best\_score}$}
            \State $\textit{best\_idx} \gets idx$
            \State $\textit{best\_score} \gets sim\_score$
        \EndIf
    \EndFor
    \State Add $\textit{best\_idx}$ to $\textit{selected\_indices}$
    \State Append $\textit{evidence\_list[best\_idx]}$ to $\textit{ranked\_evidence}$
\EndFor

\State \textbf{return} $\textit{ranked\_evidence}$
\end{algorithmic}
\end{algorithm}

\paragraph{Explanation.} 
The claim and candidate evidence sentences are first encoded into dense vector representations using the \texttt{BAAI/bge-large-en-v1} embedding model. 
The algorithm then selects evidence iteratively: at each step, it chooses the sentence that maximizes cosine similarity to the claim. 
From the second iteration onward, the similarity is computed between the claim embedding and the average embedding of the previously selected evidence combined with each remaining candidate. 
This ensures that each newly selected sentence complements the already chosen evidence to maximize relevance. 
The process continues until all evidence is ranked, producing an ordered list where sentences most relevant to the claim appear first.

\subsection{Fine-tuned NLI-based Ranking}

\begin{algorithm}[H]
\small
\caption{NLI-based Ranking of Evidence}\label{alg:nli-ranking}
\begin{algorithmic}
\State $\textit{scores} \gets \text{ComputeNLI}(\textit{claim}, \textit{evidence\_list})$ 
\State $\textit{strongest\_label} \gets \text{RetrieveAndRerank}(\textit{claim}, \textit{scores}, top\_k)$
\State $\textit{ranked\_evidence} \gets \text{SortDescending}(\textit{scores}, key=\textit{strongest\_label})$
\State \textbf{return} $\textit{ranked\_evidence}$
\end{algorithmic}
\end{algorithm}

\paragraph{Explanation.} 
In the first step, we computed the NLI score for each candidate evidence sentence with respect to the claim using the \texttt{NLI-DeBERTa-v3-large} model, producing support and refute probabilities. 
Subsequently, we applied the \textit{Retrieve-and-Rerank} procedure \citep{schuster-etal-2022-stretching} on the same NLI scores to identify which sentences exhibit stronger support for or refutation of the claim. 
Based on the strongest label determined in this step, the evidence sentences were sorted according to their corresponding NLI scores, producing a ranked list of evidence. 
Importantly, both steps rely exclusively on the same fine-tuned NLI model; the uniqueness of this approach lies in the algorithmic procedure that leverages the model in two complementary stages to achieve more effective ranking of evidence.

\begin{table}[H]
\centering
\adjustbox{max width=\linewidth}{
    \begin{tabular}{lc}
    \toprule
    \textbf{Ranking Method} & \textbf{NDCG $\uparrow$} \\
    \midrule
    One-Shot Similarity & 0.77\textsuperscript{\tiny ±0.01} \\
    Incremental Similarity & 0.79\textsuperscript{\tiny ±0.01} \\
    \midrule
    One-Shot Fine-tuned NLI & 0.57\textsuperscript{\tiny ±0.01} \\
    \midrule
    One-Shot Fine-tuned Reranker & 0.89\textsuperscript{\tiny ±0.0} \\
    \midrule
    \multicolumn{2}{l}{\textbf{GPT-4o}} \\
    One-Shot LLM & 0.91\textsuperscript{\tiny ±0.0} \\
    Incremental LLM & 0.90\textsuperscript{\tiny ±0.0} \\
    \multicolumn{2}{l}{\textbf{Qwen3-235B-A22B-Instruct}} \\
    One-Shot LLM & 0.91\textsuperscript{\tiny ±0.0} \\
    Incremental LLM & \textbf{0.92}\textsuperscript{\tiny ±0.0} \\
    \bottomrule
    \end{tabular}
}
\caption{NDCG across different ranking methods along with their standard error (SEM, see ~\ref{sec:sem_computation} for details).} 
\label{tab:ndcg-results}
\end{table}

\section{NDCG Results}
\label{sec:ndcg_results}

We also report NDCG scores for all ranking methods (Table~\ref{tab:ndcg-results}), providing a measure of ranking quality that accounts for the positions of all relevant sentences. The overall trends align with MRR and success rate: LLM-based methods and the fine-tuned reasoning-based reranker achieve the highest NDCG, indicating that they effectively prioritize the most informative sentences early in the ranking. Similarity-based methods achieve moderate NDCG values, while fine-tuned NLI-based approaches perform substantially worse. These results confirm that methods which better capture informative and complementary sentences not only improve top-k coverage but also produce higher-quality overall rankings.

\begin{figure}[H]
\centering
\includegraphics[width=\linewidth]{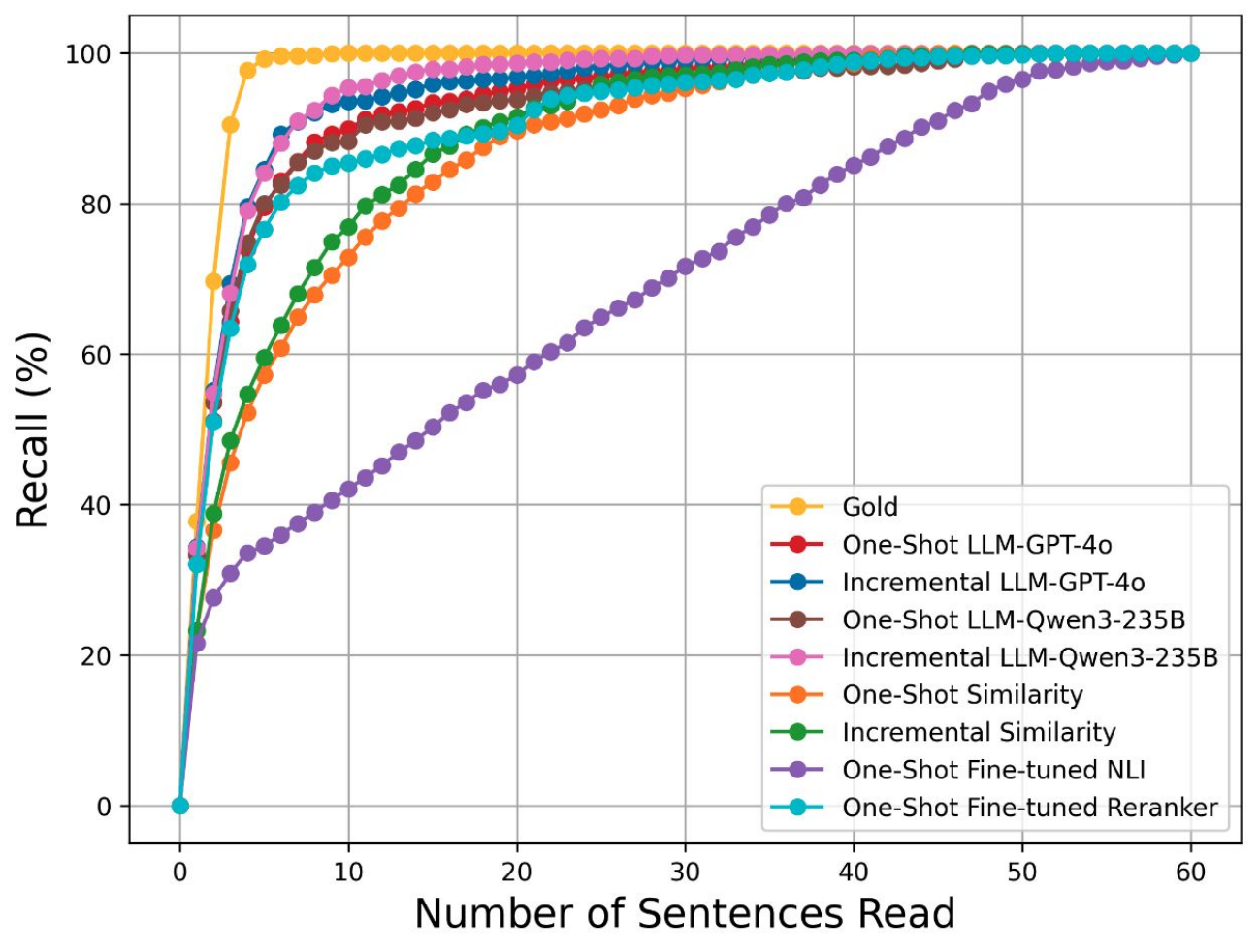}
\caption{Accumulative recall as a function of the number of sentences that the user has read. 
For LLM-based approaches, we find that the trend changes after six sentences. In other words, if the user did not find sufficient evidence in the first six sentences, they will likely have to read an excessive amount of sentences.}
\label{fig:recall_sentences_count_methods}
\end{figure}

\begin{figure}[t]
\begin{mylisting}{promptStyle}

<@\textcolor{instructionsColor}{Claim: Despicable Me 2 was directed by a lake}@>

Output:

<@\textcolor{black}{
1. ...Sentence 2: 'the film is directed by Pierre Coffin and Chris Renaud' – 'the film' refers to Despicable Me 2 from context, as it is mentioned in the sentence about Despicable Me 2. Looking back: Sentence 2 starts with 'Produced by Illumination Entertainment for Universal Pictures, and animated by Illumination Mac Guff, the film is directed...' and 'the film' is Despicable Me 2, as per the beginning. Similarly, Sentence 14 explicitly says 'Despicable Me 2'. Sentence 14: 'including Despicable Me and Despicable Me 2, which he co-directed with Pierre Coffin'. Sentence 17 says 'the films in the Despicable Me franchise', which includes Despicable Me 2. ...}@>

<@\textcolor{black}{
2. ...But let's see Sentence 14: 'Despicable Me 2, which he co-directed with Pierre Coffin' – 'he' refers to Chris Renaud, so it is explicit. Similarly, Sentence 17: 'co-directing, with Chris Renaud, the films' – so Pierre Coffin co-directed. But for Despicable Me 2 specifically, Sentence 14 mentions it by name. ...}@>

<@\textcolor{black}{
3. ...Sentence 12: 'Chris Renaud -LRB- born December 1966 -RRB- is an American illustrator and filmmaker.' ...}@>

\end{mylisting}
\caption{Example output of Qwen3-235B-Thinking.}
\label{fig:qwen_example}
\end{figure}

\section{Analysis of Qwen3-235B-Thinking Behavior}
\label{sec:qwen3_output_example}
The Qwen3-235B-Thinking model often provides detailed per-sentence reasoning, but does not produce a valid selection of the next sentence.
In such cases, we defaulted to the textual order of sentences (i.e. their reading sequence), which naturally limited the observed performance. We experimented with different prompt variations and techniques, including structured output formats to enforce ranking selection. While structured output helped the model consistently produce a selection, it ultimately resulted in worse overall performance; therefore, we report results using the prompt with detailed instructions that balances reasoning and ranking accuracy. 

For example, see \cref{fig:qwen_example}. These citations collectively illustrate the behavior of Qwen3-235B-Thinking. The model produces thorough, step-by-step reasoning across multiple sentences but does not consistently select the most relevant sentence at each step. For instance, while sentences 2, 14, and 17 provide key evidence about Despicable Me 2 and its directors, the model spreads its attention across them without prioritizing the most critical one. Sentence 12 is necessary to verify that Chris Renaud is an American filmmaker, yet the model does not explicitly link it to the claim, leaving some aspects unresolved. As a result, although the model demonstrates strong per-sentence reasoning, it does not produce a clear next-step selection, which complicates ranking and can lead to missing crucial information.

\section{Analysis of Incremental NLI Performance}
\label{sec:incremental_nli_analysis}
We observed that the incremental NLI strategy underperforms the non-incremental approach. This is primarily due to the limitations of \texttt{NLI-DeBERTa-v3-large}, which is trained on single-premise, single-hypothesis pairs and is not designed to handle concatenated multi-sentence input. When sentences are concatenated, the model evaluates unnatural, multi-sentence inputs, which can reduce its ability to accurately assess the relevance of each sentence for verification. As a result, the incremental NLI approach achieves slightly lower performance than the non-incremental method, with MRR of 0.29 (vs. 0.31), Success Rate of 22.3\% (vs. 22.5\%), and NDCG of 0.55 (vs. 0.57). In practice, this leads to mis-ranking of important sentences and lower overall effectiveness. This example highlights why incremental selection, while effective for LLM-based methods, is less suitable for NLI models due to architectural and training constraints.

\section{Detailed Qualitative Analysis}
\label{sec:detailed_qualitative_analysis}

Figure~\ref{fig:qualitative_example} illustrates the ranking sequence for the example discussed in \cref{sec:qalitive_analysis_discussion}. The incremental LLM-based method reached gold coverage after reading 4 sentences, while the one-shot LLM required 5. The fine-tuned reasoning-based reranker also required 4 sentences to reach sufficient evidence. For similarity-based methods, incremental similarity needed 6 sentences and one-shot similarity 8 sentences to cover sufficient evidence. Our analysis revealed that, in addition to relevant sentences, highly ranked sentences often included ones not part of any gold set; these were topically related to the claim but lacked the information necessary to support or refute it, which accounts for their high ranking. The fine-tuned NLI approach required 14 sentences in total.

\begin{figure*}[p]
\begin{mylisting}{promptStyle}
<@\textcolor{instructionsColor}{Claim:}@>
<@\textcolor{black}{One of his tasks was to prepare the ground for the reintroduction of compulsory national service; the new scheme was enacted in 1951 and remained in force until 1959.}@>

<@\textcolor{instructionsColor}{Gold verdict:}@> supported.

<@\textcolor{instructionsColor}{Gold optimal evidence sets:}@>
<@\textcolor{black}{Set 1:}@>
<@\textcolor{black}{[9] In the context of the intensification of the Cold War in Europe, Communist insurgency and success in South-East Asia, and the declaration of war in Korea, the Menzies government sponsored the National Service Act 1951.}@>
<@\textcolor{black}{[18] On 24 November 1959 Cabinet decided that National Service call-ups should be terminated and that arrangements for the January 1960 intake would be cancelled.}@>

<@\textcolor{black}{Set 2:}@>
<@\textcolor{black}{[14] The first call-up notice was issued on 12 April 1951.}@>
<@\textcolor{black}{[18] On 24 November 1959 Cabinet decided that National Service call-ups should be terminated and that arrangements for the January 1960 intake would be cancelled.}@>

<@\textcolor{instructionsColor}{All evidence sets:}@>
<@\textcolor{black}{1. [9,18]}@>
<@\textcolor{black}{2. [14,18]}@>
<@\textcolor{black}{3. [1,10,14]}@>

<@\textcolor{red}{Output rankings provided by the different methods:}@>
LLM: [<@\textcolor{instructionsColor}{9}@>,<@\textcolor{darkyellow}{10}@>,<@\textcolor{black}{15}@>,<@\textcolor{darkyellow}{14}@>,<@\textcolor{instructionsColor}{18}@>,...]
Incremental LLM: [<@\textcolor{instructionsColor}{9}@>,<@\textcolor{darkyellow}{14}@>,<@\textcolor{darkyellow}{10}@>,<@\textcolor{instructionsColor}{18}@>,...]
Fine-tuned Reasoning-Based Reranker: [<@\textcolor{instructionsColor}{9}@>,<@\textcolor{darkyellow}{10}@>,<@\textcolor{black}{15}@>,<@\textcolor{instructionsColor}{18}@>,...]
Similarity: [<@\textcolor{black}{20}@>,<@\textcolor{instructionsColor}{18}@>,<@\textcolor{black}{8}@>,<@\textcolor{darkyellow}{1}@>,<@\textcolor{darkyellow}{10}@>,<@\textcolor{black}{19}@>,<@\textcolor{black}{15}@>,<@\textcolor{instructionsColor}{14}@>,...]
Incremental Similarity: [<@\textcolor{black}{20}@>,<@\textcolor{darkyellow}{10}@>,<@\textcolor{black}{8}@>,<@\textcolor{instructionsColor}{18}@>,<@\textcolor{black}{12}@>,<@\textcolor{instructionsColor}{9}@>,...]
Fine-tuned NLI: [<@\textcolor{black}{16}@>,<@\textcolor{instructionsColor}{18}@>,<@\textcolor{darkyellow}{10}@>,<@\textcolor{black}{12}@>,<@\textcolor{black}{32}@>,<@\textcolor{black}{29}@>,<@\textcolor{black}{25}@>,<@\textcolor{black}{13}@>,<@\textcolor{black}{31}@>,<@\textcolor{black}{7}@>,<@\textcolor{black}{11}@>,<@\textcolor{black}{19}@>,<@\textcolor{black}{17}@>,<@\textcolor{instructionsColor}{14}@>,...]

\end{mylisting}
\caption{Illustrative example from the dataset showing ranked sentences for each method until one gold set is fully covered. Sentences in green belong to the covered gold set, yellow indicates sentences from other gold sets not yet covered, and black marks sentences that are not part of any gold set.}
\label{fig:qualitative_example}
\end{figure*}

\section{Standard Error of the Mean (SEM) Calculation}
\label{sec:sem_computation}

For each ranking method, we report the standard error of the mean (SEM) to indicate variability across instances. SEM is computed as:

\[
\text{SEM} = \frac{\sigma}{\sqrt{n}}
\]

where \(\sigma\) is the standard deviation of the metric values across all instances, and \(n\) is the number of instances in the dataset. This provides an estimate of the uncertainty of the reported mean performance metrics (MRR, SR and NDCG) for each method.

\section{User Study Details}
\label{app:user_study_detailed}

\subsection{Participants}
Five fluent English speakers with a computer-science background participated in the study. All participants provided informed consent.

\subsection{Materials and Design}
We used 100 claims from our curated dataset. To avoid carryover effects, the claims were partitioned into five disjoint subsets; each participant saw 40 unique trials (20 per condition). No claim appeared twice for any participant. Each trial consisted of a claim paired with evidence produced by one of the evaluated systems:

\begin{itemize}
    \item \textbf{Evidence Ranking:} Incremental LLM evidence ranking with GPT-4o.
    \item \textbf{LLM Evidence Selection:} Sentence-level evidence selection method adapted from \citep{hirsch-etal-2025-laquer}.
\end{itemize}

\subsection{Procedure}
A web-based interface presented one claim per trial, with the interaction logic determined by the experimental condition. 

\begin{figure*}[t]
    \centering
    \includegraphics[width=0.95\textwidth]{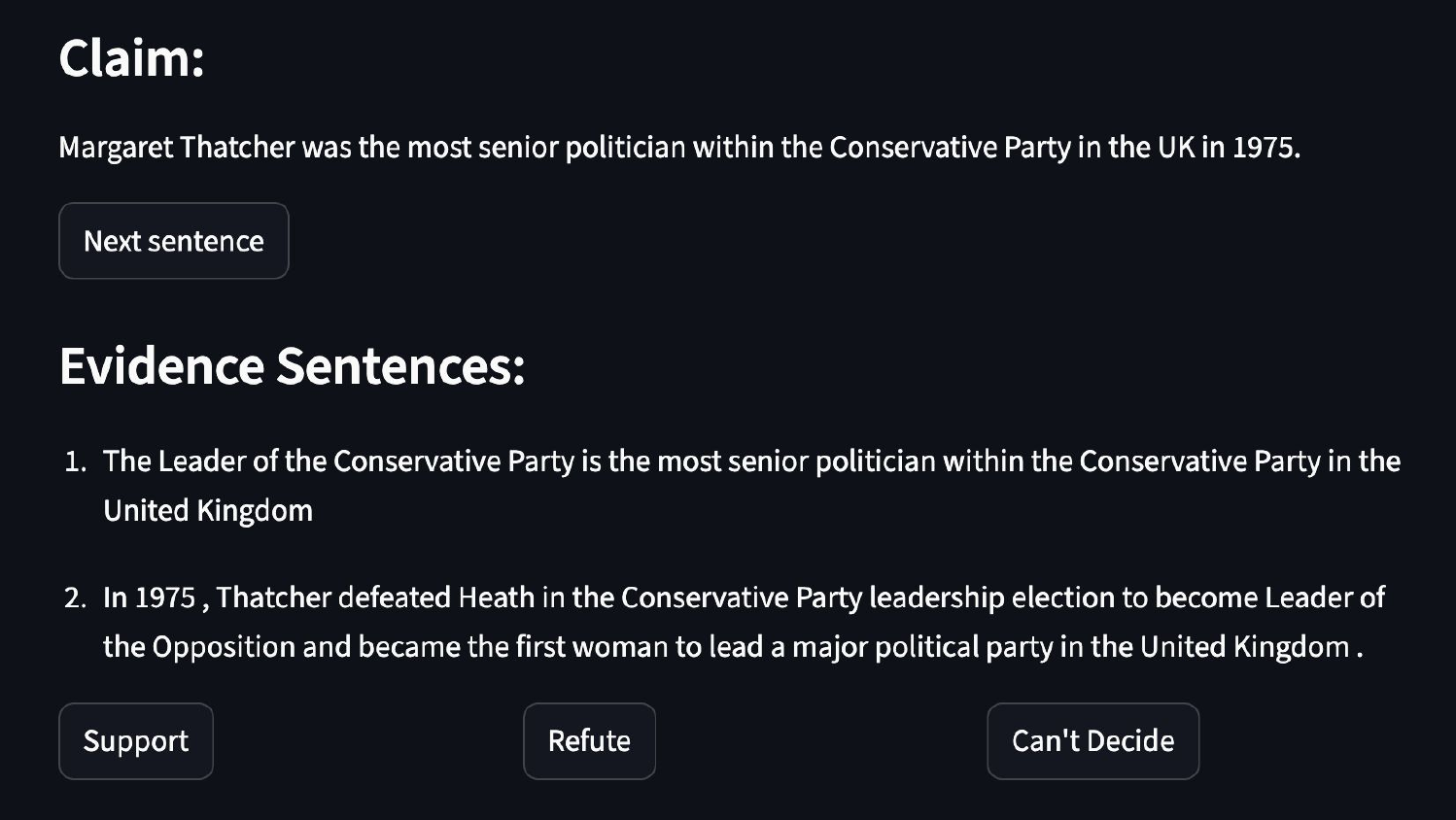}
    \caption{\textbf{Evidence Ranking Condition Interface.} The interface allows users to reveal evidence incrementally, supporting an active search and verification process.}
    \label{fig:ranking_example}
\end{figure*}

\textbf{Evidence Ranking Condition:} As shown in Figure~\ref{fig:ranking_example}, participants initially viewed only the top-ranked sentence. They could incrementally reveal additional evidence by clicking ``show next sentence'' once they judged the current information insufficient to reach a decision.

\begin{figure*}[t]
    \centering
    \includegraphics[width=0.95\textwidth]{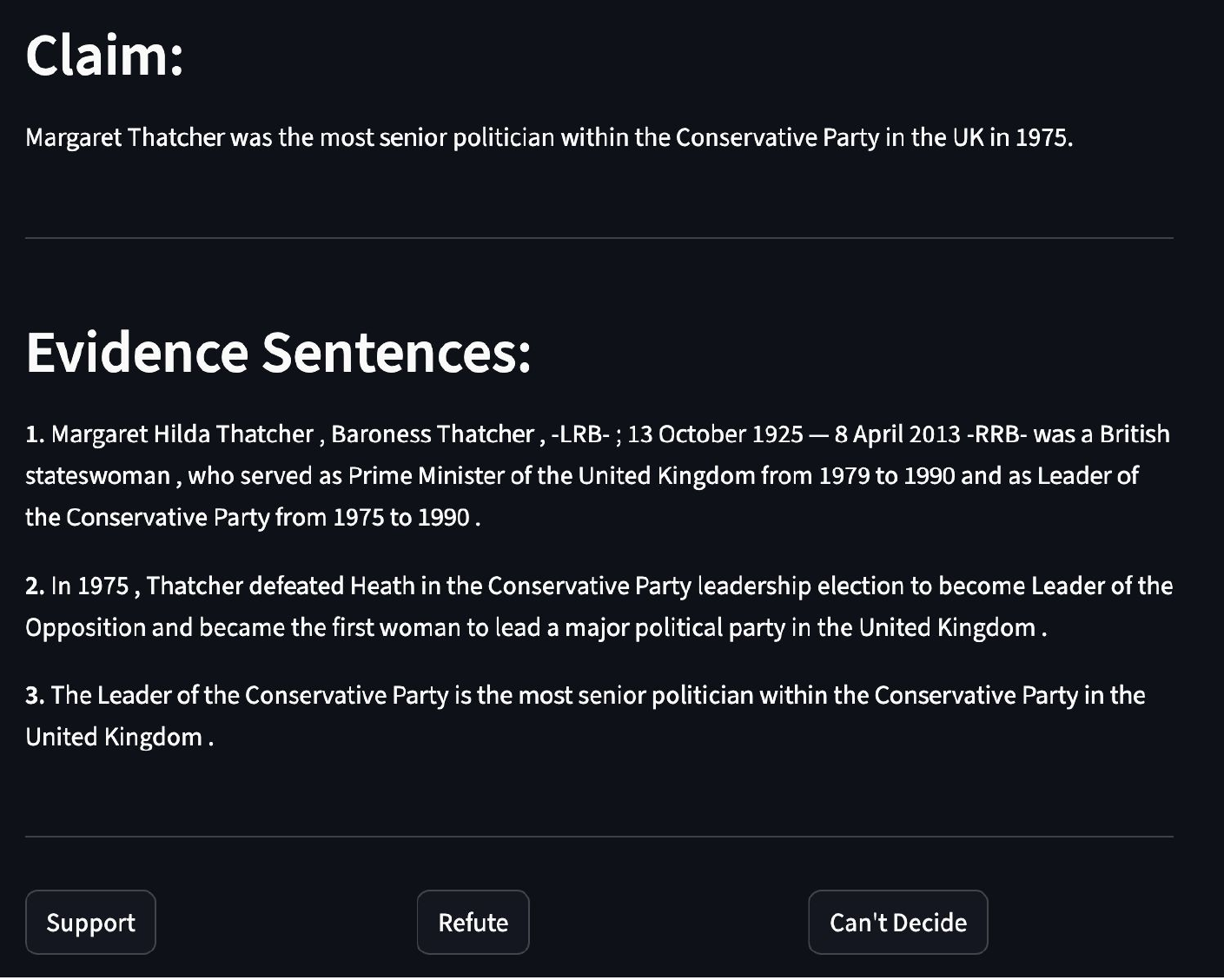}
    \caption{\textbf{LAQuer Condition Interface.} The interface presents all system-selected sentences at once, allowing for immediate global assessment of the claim's validity.}
    \label{fig:laquer_example}
\end{figure*}

\textbf{Evidence Selection Condition:} As shown in Figure~\ref{fig:laquer_example}, participants viewed all system-selected sentences simultaneously. They had no control over the disclosure of information and were presented with the full set of evidence at once.

\subsection{Measures}
For each trial, we recorded: (1) participant's final decision, (2) whether the decision matched the gold label, and (3) the number of evidence sentences read.

These measures capture both verification accuracy and information efficiency, enabling analysis of user reading effort and model performance in practical verification scenarios.

\end{document}